\documentclass{article} 

     \usepackage[final,nonatbib]{neurips_2021}

\usepackage{amsmath,amsfonts,bm}

\def\eqref#1{equation~\ref{#1}}

\def\1{\bm{1}}

\def\eps{{\epsilon}}

\DeclareMathAlphabet{\mathsfit}{\encodingdefault}{\sfdefault}{m}{sl}
\SetMathAlphabet{\mathsfit}{bold}{\encodingdefault}{\sfdefault}{bx}{n}

\DeclareMathOperator*{\argmax}{arg\,max}

\usepackage[dvipsnames]{xcolor}
\usepackage{hyperref}
\usepackage{url}
\usepackage[T1]{fontenc}
\usepackage{booktabs} 
\usepackage{footnote}
\usepackage{graphicx}
\usepackage{amsthm}
\usepackage{algorithm}
\usepackage[noend]{algpseudocode}
\usepackage{wrapfig}

\usepackage[normalem]{ulem}

\title{No-Press Diplomacy from Scratch}

\author{
 Anton Bakhtin \enskip David Wu \enskip Adam Lerer  \enskip Noam Brown \\
 Facebook AI Research  \\
  \texttt{\{yolo,dwu,alerer,noambrown\}@fb.com}  \\
}

\newcommand{\branchingFactor}{$10^{20}$}
\newcommand{\botname}{\emph{DORA}}
\newcommand{\botFvA}{\botname}
\newcommand{\botHumanInit}{\emph{HumanDNVI-NPU}}
\newcommand{\botHumanInitWild}{\emph{HumanDNVI}}
\newcommand{\scratchDNLBot}{\emph{ScratchDNVI}}

\newcommand{\myvector}[1]{\bm{#1}}

\begin{document}

\maketitle

\begin{abstract}
Prior AI successes in complex games have largely focused on settings with at most hundreds of actions at each decision point.
In contrast, Diplomacy is a game with more than \branchingFactor{} possible actions per turn.
Previous attempts to address games with large branching factors, such as Diplomacy, StarCraft, and Dota, used human data to bootstrap the policy or used handcrafted reward shaping. 
In this paper, we describe an algorithm for action exploration and equilibrium approximation in games with combinatorial action spaces. This algorithm simultaneously performs value iteration while learning a policy proposal network.
A double oracle step is used to explore additional actions to add to the policy proposals. At each state, the target state value and policy for the model training are computed via an equilibrium search procedure. 
Using this algorithm, we train an agent, \botFvA{}, completely from scratch for a popular two-player variant of Diplomacy and show that it achieves superhuman performance. Additionally, we extend our methods to full-scale no-press Diplomacy and for the first time train an agent from scratch with no human data.
We present evidence that this agent plays a strategy that is incompatible with human-data bootstrapped agents. This presents the first strong evidence of multiple equilibria in Diplomacy and suggests that self play alone may be insufficient for achieving superhuman performance in Diplomacy.
\end{abstract}

\section{Introduction}

\begin{wrapfigure}{r}{0.37\textwidth}
    \includegraphics[width=1.01\linewidth]{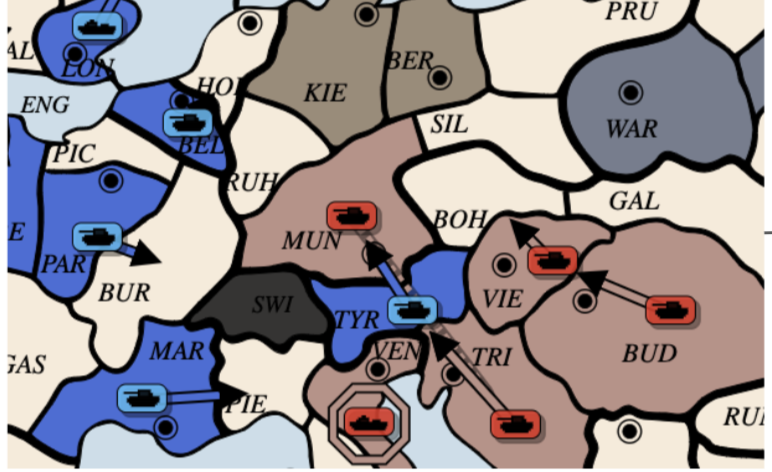}

   \caption{France (blue) would be in a strong position were it not for exactly two actions that Austria (red) can play to dislodge the unit in Tyr. The probability of sampling one of them randomly is roughly $10^{-6}$.}
   \label{fig:teaser}
\end{wrapfigure}

Classic multi-agent AI research domains such as chess, Go, and poker feature action spaces with at most thousands of actions per state, and are therefore possible to explore exhaustively a few steps deep. In contrast, modern multi-agent AI benchmarks such as StarCraft, Dota, and Diplomacy feature combinatorial action spaces that are incredibly large. In StarCraft and Diplomacy, this challenge has so far been addressed by bootstrapping from human data~\cite{vinyals2019grandmaster,paquette2019no,anthony2020learning,gray2020human}. In Dota, this challenge was addressed through careful expert-designed reward shaping~\cite{berner2019dota}.

In contrast, in this paper we describe an algorithm that trains agents through self play without any human data and can accommodate the large action space of Diplomacy, a long-standing AI benchmark game in which the number of legal actions for a player on a turn is often more than \branchingFactor{}~\cite{paquette2019no}.

We propose a form of deep RL (reinforcement learning) where at each visited state the agent approximates a Nash equilibrium for the stage game with 1-step lookahead search, plays the equilibrium policy, and uses the equilibrium value as the training target. Since the full action space is too large to consider, this equilibrium is computed over a smaller set of candidate actions. These actions are sampled from an action proposal network that is trained in unison with the value model. We show that our deep Nash value iteration algorithm, when initialized with a human-bootstrapped model, defeats multiple prior human-derived agents in no-press Diplomacy by a wide margin.

Next, we investigate training an agent completely from scratch with no human data in Diplomacy. In order to improve action exploration when training from scratch, the proposal actions are augmented with exploratory actions chosen through a double-oracle-like process, and if these discovered actions end up as part of the equilibrium then they will be learned by the proposal network.
Using this technique, which we call \botFvA{} (Double Oracle Reinforcement learning for Action exploration), we develop a superhuman agent for a popular two-player variant of Diplomacy. Our agent is trained purely through self-play with no human data and no reward shaping.

We also train a \botname{} agent for the 7-player game of no-press Diplomacy,
the first no-press Diplomacy agent to be trained entirely from scratch with no human data.
When 6 \botname{} agents play with 1 human-data-based agent, all human-data-based agents perform extremely poorly. However, when one \botname{} agent plays with 6 human-data-based agents, the bot underperforms.
These results suggest that self-play in 7-player Diplomacy may converge to equilibria that are incompatible with typical human play. This is in contrast to previous large-scale multi-agent benchmarks like multiplayer poker~\cite{brown2019superhuman} and validates Diplomacy as a valuable benchmark for learning policies that are compatible with human behavior.
\section{Background and Related Work}

\subsection{Description of Diplomacy}
In this section we summarize the rules of Diplomacy. For a more detailed description, see~\cite{paquette2019no}.
No-press Diplomacy is a zero-sum board game where 7 players compete for control on a map of Europe. The map contains 75 locations, 34 of which are \emph{supply centers} (SCs).
Each player begins with 3-4 units and 3-4 SCs and a player wins by controlling a majority (18) of the SCs. If no player controls a majority and all remaining players agree to a draw, then the victory is divided between the uneliminated players.
In case of a draw, we use the \textbf{Sum-of-Squares (SoS)} scoring system that defines the score of player~$i$ as $C_i^2 /\sum_{i'}C_{i'}^2$, where $C_i$ is the SC count for player~$i$.

Each turn during movement, all players simultaneously choose an action composed of one order for every unit they control. An order may direct a unit to hold stationary, to move to an adjacent location, to support a nearby unit's hold or move, to move across water via convoy, or to act as the convoy for another unit. Because a player may control up to 17 units with an average of 26 valid orders for each unit, the number of possible actions is typically too large to enumerate. 
Additionally, in Diplomacy, often supports or other critical orders may be indistinguishable from a no-op except in precise combinations with other orders or opponent actions. This makes the action space challenging to learn with black-box RL. Actions are simultaneous, so players typically must play a mixed strategy to avoid being predictable and exploitable.


In the popular \emph{France vs. Austria (FvA)} variant of Diplomacy, the number of players is reduced to two, but the victory condition and all game rules, including the number of locations and centers, remain the same. FvA serves as an effective way to isolate the challenge of an astronomically large imperfect-information action space from the challenge of Diplomacy's multi-agent nature.

\subsection{Stochastic Games}
\emph{Stochastic games}~\cite{shapley1953stochastic} are a multi-agent generalization of Markov decision processes. At each state, each player chooses an action and the players' joint action determines the transition to the next state.
Formally, at each state $s \in S$ agents simultaneously choose a joint action $\myvector{a} = (a_1, \dots, a_\mathcal{N}) \in A_1 \times \dots \times A_{\mathcal{N}}$, where $\mathcal{N}$ is the number of players.
Agents then receive a reward according to a vector function $\myvector{r}(s, \myvector{a}) \in \mathbb{R}^\mathcal{N}$ and transition to a new state $s' = f(s, \myvector{a})$.
The game starts at initial state $s^0$ and continues until a terminal state is reached, tracing a trajectory $\tau = ((s^0,\myvector{a}^0),...,(s^{t-1},\myvector{a}^{t-1}))$ where $f(s^{t-1},\myvector{a}^{t-1})$ is terminal. Each agent~$i$ plays according to a policy $\pi_i: S \rightarrow \Delta A_i$, where $\Delta A_i$ is a set of probability distributions over actions in $A_i$.
Given a policy profile $\myvector{\pi} = (\pi_1, \dots, \pi_\mathcal{N})$, the probability of a trajectory is $p(\tau|\myvector{\pi}) = \prod_{(s,\myvector{a}) \in \tau} p(\myvector{a} | \, \myvector{a} \sim \myvector{\pi}(s))$, and we define the vector of expected values of the game rooted at state $s$ as $\myvector{V}(\myvector{\pi}, s) = \sum_{\tau \in T(s)} p(\tau|\myvector{\pi}) \sum_{(s_t,\myvector{a_t}) \in \tau} \gamma^t \myvector{r}(s_t, \myvector{a_t})$ where $T(s)$ is the set of terminal trajectories beginning at $s$.
We denote by $\myvector{V}(\myvector{\pi}) := \myvector{V}(\myvector{\pi}, s^0)$ the value of a game rooted in the initial state. We denote by $\myvector{\pi}_{-i}$ the policies of all players other than~$i$.



A \textbf{best response (BR)} for agent~$i$ against policy profile $\pi$ is any policy that achieves maximal expected value. That is, $\pi_i^*$ is a BR to $\pi$ if $V_i(\pi_i^*, \pi_{-i}) = \max_{\pi'_i} V_i(\pi'_i, \pi_{-i})$. A policy profile is a \textbf{Nash equilibrium (NE)} if all players' policies are BRs to each other. Letting $V_i$ be the $i$th component of $\myvector{V}$, $\myvector{\pi}$ is a NE if:
\begin{equation}
\forall i \, \forall \pi'_i : V_i(\myvector{\pi}) \geq V_i(\pi'_i, \myvector{\pi}_{-i})
\nonumber
\end{equation}
An \textbf{$\epsilon$-NE} is a policy profile in which an agent can only improve its expected reward by at most $\epsilon$ by switching to a different policy. If a policy profile is an $\epsilon$-NE, we say its \textbf{exploitability} is $\epsilon$.

We may also talk about Nash equilibria for one stage of the game rather than the whole game. For any state $s$, supposing we have values $Q(s,\myvector{a})$ that represent the expected values of playing joint action $\myvector{a}$ in state $s$ assuming some policy profile is played thereafter, we may define a new 1-step game where instead the game ends immediately after this action and $Q(s,\myvector{a})$ is itself the final reward. For such stage games, notationally we denote NE using the symbol $\sigma$ rather than $\pi$.
\subsection{Nash Q-Learning}
\label{section:nashqlearning}
Value iteration methods such as Nash-Q have been developed that provably converge to Nash equilibria in some classes of stochastic game, including 2p0s (two-player zero-sum) stochastic games~\cite{littman1994markov,hu2003nash}. Nash-Q learns joint Q values $\myvector{Q}(s, \myvector{a})$ that aim to converge to the state-action value of $(s, \myvector{a})$ assuming that some NE $\myvector{\pi}$ is played thereafter.
This is done by performing 1-step updates on a current estimated function $\myvector{Q}$ as in standard Q-learning, but replacing the max operation with a stage game NE computation.

Formally, suppose that $\myvector{a}$ is played in $s$ leading to $s'$. A NE~${\sigma}$, $\sigma_i \in \Delta A_i$, is computed for the stage game at $s'$ whose payoffs for each joint action $\myvector{a}'$ are specified by the current values of $\myvector{Q}(s',\myvector{a}')$. The value of this game to each player assuming this NE is then used as the target values of $\myvector{Q}(s, \myvector{a})$. The update rule is thus:
\begin{equation}\label{eq:nashq}
\myvector{Q}(s, \myvector{a}) \leftarrow (1 - \alpha) \myvector{Q}(s, \myvector{a}) + \alpha \Big(\myvector{r}(s, \myvector{a}) + \gamma \sum\limits_{\myvector{a}'} \sigma(\myvector{a'}) \myvector{Q}(s', \myvector{a}')\Big),
\end{equation}
where $\alpha$ is the learning rate, $\gamma$ is the discount factor, and ${\sigma}(\myvector{a}) := \prod_i\sigma_i({a_i})$ is the probability of joint action $\myvector{a}$.


As presented later in Section \ref{section:deepnashlearning}, the foundation of our algorithm is an adaptation of Nash Q-Learning, with several modifications to make it suitable for deep RL and adjusted to use a value function rather than a Q function.



\subsection{Double Oracle}
\label{section:doubleoracle}
\textbf{Double oracle (DO)} is an iterative method for finding a NE over a large set of actions in a matrix game \cite{mcmahan2003doubleoracle}. It leverages the fact that computing a best response is computationally much cheaper than computing a NE. Starting from a small set of candidate actions $A^0_i \subseteq A_i$ for each player~$i$, on each iteration it computes a NE~${\sigma}^t$ in the matrix game restricted to the candidate actions $\myvector{A}^t$. Then, at each iteration~$t$ for each player~$i$ it finds the best response action~$a^{t + 1}_i \in A_i$ to the restricted NE~$\sigma^t$ and creates ${A}^{t + 1}_i := {A}^t_i \bigcup \{a^{t+1}_i\}$. Formally, $a^{t+1}_i$ is an action in the full action set $A_i$ that satisfies
\begin{equation}
a^{t + 1}_i = \argmax_{a'\in A_i} \sum_{\myvector{a} \in \myvector{A}^t} Q_i(s,{a}_1,\dots,{a}_{i-1},a',{a}_{i+ 1},\dots,{a}_\mathcal{N}) \prod_{j\neq i} \sigma^t_j(a_j).
\end{equation}
DO-inspired methods have been used in a variety of other work for handling large or continuous action spaces \cite{jain2011double,brown2015simultaneous,lanctot2017psro,lukas2020continuous,mcaleer2020pipeline}. Although in the worst case the equilibrium policy may have support over a large fraction of the action space, in practice in many problems a small subset of actions may suffice, and in this case DO-based methods can be computationally efficient due to being able to restrict equilibrium computation to only a tiny fraction of the total number of actions.

As we will present in Section \ref{section:actionexploration}, we make use of a DO-like procedure both within the RL loop and at inference time, and show it to be an effective way to introduce novel good actions to the policy.

\subsection{Regret Matching}
\label{section:sampledregretmatching}
Both Nash-Q-Learning and DO require the computation of NEs in matrix games, which we approximate using the iterative \textbf{regret matching (RM)} algorithm~\cite{blackwell1956analog,hart2000simple}, as has been done in past work~\cite{gray2020human}. RM is only guaranteed to converge to a NE in 2p0s games and other special classes of games~\cite{hannan1957approximation}. Nevertheless, past work has found that it produces competitive policies that have low exploitability in important non-2p0s games including six-player poker and no-press Diplomacy~\cite{brown2019superhuman,gray2020human}. We therefore use RM to approximate a NE~$\sigma$ in restricted matrix games during DO.

In RM, each player~$i$ has a \textbf{regret} value for each action~$a_i \in {A}_i$. The regret on iteration~$t$ is denoted $\text{Regret}_i^t(a_i)$. Initially, all regrets are zero. On each iteration~$t$ of RM, $\sigma^t_i(a_i)$ is set according to
\begin{equation} \label{eq:rm}
    \sigma^{t}_i(a_i) =
    \begin{cases}
        \frac{\max\{0,\text{Regret}^t_i(a_i)\}}{\sum_{a'_i \in \mathcal{A}_i} \max\{0,\text{Regret}^t_i(a'_i)\}}  & \text{if } \sum_{a'_i \in \mathcal{A}_i} \max\{0,\text{Regret}^t_i(a'_i)\} > 0 \\
        \frac{1}{|\mathcal{A}_i|}                                     & \text{otherwise} \\
    \end{cases}
\end{equation}
Next, each player samples an action $a^*_i$ from $\mathcal{A}_i$ according to $\sigma_i^t$ and all regrets are updated such that
\begin{equation} \label{eq:srm}
\text{Regret}_i^{t+1}(a_i) = \text{Regret}_i^{t}(a_i) + v_i(a_i, a^*_{-i}) - \sum_{a'_i \in \mathcal{A}_i} \sigma^t_i(a'_i) v_i(a'_i, a^*_{-i}),
\end{equation}
where $v_i(\overline{a})$ is the utility of player $i$ in the matrix game.
This sampled form of RM guarantees that the \emph{average} policy over all iterations converges to a $\epsilon$-NE in 2p0s games at a rate of at worst $\mathcal{O}(\frac{1}{\epsilon^2})$ iterations with high probability~\cite{lanctot2009monte}.
In order to improve empirical performance, we use linear RM~\cite{brown2019solving}, which weighs updates on iteration~$t$ by~$t$.\footnote{In practice, rather than weigh iteration~$t$'s updates by $t$ we instead discount prior iterations by $\frac{t}{t+1}$ in order to reduce numerical instability. The two options are mathematically equivalent.} We also use optimism~\cite{syrgkanis2015fast}, in which the most recent iteration is counted twice when computing regret.



\section{Algorithms}
We now describe DORA in detail. DORA simultaneously learns a state-value function and an action proposal distribution via neural networks trained by bootstrapping on an approximate Nash equilibrium for the stage game each turn. A DO-based process is used to discover actions. 


\subsection{Deep Nash Value Iteration}
\label{section:deepnashlearning}
The core of the learning procedure is based on Nash Q-Learning as presented in Section \ref{section:nashqlearning}, but simplified to use only a value function, with adaptations for large action spaces and function approximation.

Approximating a Q-function in a game with an action space as large as Diplomacy is challenging. However, in Diplomacy as in many other domains, the reward $\myvector{r}(s,a)$ depends only on the next state $s' = f(s,a)$ and not on how it was reached, 
and since the transition function is known and can be simulated exactly, we can redefine $\myvector{r}$ to be a function of just the next state $s'$. In that case, we can rewrite Update Rule~\ref{eq:nashq} (from Section \ref{section:nashqlearning}) in terms of a state value function:
\begin{equation}\label{eq:nashv}
\myvector{V}(s) \leftarrow (1 - \alpha) \myvector{V}(s) + \alpha (\myvector{r}(s) + \gamma \sum\limits_{\myvector{a'}} \sigma(\myvector{a'}) \myvector{V}(f(s, \myvector{a'})))
\end{equation}
Since the state space is large, we use a deep neural network $\myvector{V}(s; \theta_v)$ with parameters $\theta_v$ to approximate $\myvector{V}(s)$. Since the action space is large, exactly computing the Nash equilibrium $\sigma$ for the 1-step matrix game at a state is also infeasible. Therefore, we train a policy proposal network $\pi(s;\theta_\pi)$ with parameters $\theta_\pi$ that approximates the distribution of actions under the equilibrium policy at state $s$. We generate a candidate action set at state $s$ by sampling $N_{b}$ actions from $\pi(s;\theta_\pi)$ for each player and selecting the $N_{c}\ll N_{b}$ actions with highest likelihood. We then approximate a NE $\sigma$ via RM (described in Section~\ref{section:sampledregretmatching}) in the restricted matrix game in which each player is limited to their $N_{c}$ actions assuming successor states' values are given by our current value network $\myvector{V}(s;\theta_v)$.

We explore states via self-play to generate data, with both players following the computed NE policy with an $\epsilon$ probability of instead exploring a random action from among the $N_c$ proposals. Also, in place of $\theta_{v}$ and $\theta_\pi$, self-play uses slightly-lagged versions of the networks $\hat{\theta}_{v}$ and $\hat{\theta}_{\pi}$ to compute $\sigma$ that are only updated periodically from the latest $\theta_{v}$ and $\theta_\pi$ in training, a normal practice that improves training stability~\cite{mnih2015human}. We then regress the value network towards the computed stage game value under the NE $\sigma$ using MSE loss (whose gradient step is simply the above update rule), and regress the policy proposal network towards $\sigma$ using cross entropy loss:
\begin{equation}\label{eq:loss}
\begin{aligned}
&\text{ValueLoss}(\theta_v) = \frac{1}{2}\left(\myvector{V}(s;\theta_v) - \myvector{r}(s) - \gamma\sum\limits_{\myvector{a'}} \sigma(\myvector{a'}) \myvector{V}\left(f(s, \myvector{a'});\hat{\theta}_v\right)\right)^2\\
&\text{PolicyLoss}(\theta_\pi) = - \sum_i \sum_{a_i \in A_i} \sigma_i(a)\log\pi_i(s, a_i; \theta_{{\pi}}),
\end{aligned}
\end{equation}
We show in Appendix~\ref{app:convergence} that in 2p0s games, the exact tabular form of the above algorithm with mild assumptions and without the various approximations for deep RL provably converges to a NE. Outside of 2p0s games and other special classes, both Nash Q-learning and RM lack theoretical guarantees; however, prior work on Diplomacy and 6-player poker have shown that similar approaches still often perform well in practice \cite{gray2020human,anthony2020learning,brown2019superhuman}, and we also observe that DORA performs well in 7-player no-press Diplomacy.


At a high level our algorithm is similar to previous algorithms such as AlphaZero~\cite{silver2018general} and ReBeL~\cite{brown2020combining} that update both a policy and value network based on self-play, leveraging a simulator of the environment to perform search.
The biggest differences between our algorithm and AlphaZero are that 1-ply RM acts as the search algorithm instead of Monte Carlo tree search (just as in~\cite{gray2020human}), the value update is uses a 1-step bootstrap instead of the end-of-game-value (which allows the trajectory generation to be off-policy if desired), and the policy network acts as an action proposer for the search but does not regularize the search thereafter. Our algorithm is also similar to a recent method to handle large action spaces for Q-learning~\cite{van2020q} and reduces to it in single-agent settings.

\subsection{Double-Oracle-Based Action Discovery}
\label{section:actionexploration}

The algorithm in Section~\ref{section:deepnashlearning} relies on the policy network to generate good candidate actions. However, good actions initially assigned too low of a probability by the network may never be sampled and reinforced.
To address this, we use a mechanism based on double oracle, described in Section~\ref{section:doubleoracle}, to introduce novel actions.
However, vanilla DO requires computing the expected value of all actions against the restricted matrix game NE. Since our action space is too large to enumerate, we generate only a pool of $N_{p} \gg N_c$ actions to consider.

A natural idea to generate this pool would be to sample actions uniformly at random from all actions. However, our results in Section \ref{sec:resultsfva} show this approach performs poorly. Intuitively, in a combinatorial action space as large as in Diplomacy often only a minuscule fraction of actions are reasonable, so random sampling is exceedingly unlikely to find useful actions (see Fig~\ref{fig:teaser} for an example).

We instead generate the pool of $N_p$ actions for a player by generating local modifications of the candidate actions already in the restricted matrix game equilibrium. We first sample uniformly among the player's $N_d$ ($d \leq c$) most probable actions in that equilibrium. Next, for each sampled action, we randomly sample a single location on the map and consider all units of the player adjacent to the location. For those units, we randomly choose legal valid new orders. We add the order in this pool that best-responds to the equilibrium as a new candidate action, and DO proceeds as normal.




As shown in Table~\ref{tab:selfplay_do}, we found this method of action sampling to outperform simply picking legal and valid combinations of orders uniformly at random.
Intuitively, this is likely because locality in Diplomacy, as in a wide variety of other problem domains and algorithms, is a powerful inductive bias and heuristic. Perturbations of good actions are vastly more likely than random actions to also be good, and spatially-nearby actions are more likely to need to covary than spatially-distant ones.


To further improve the scalability of our algorithm, we apply several additional minor approximations, including computing some of the above expected values using sampling and capping the number of DO iterations we perform. For details, see Appendix \ref{app:params}.

\section{Implementation and Methods}
We describe our implementation of the above algorithms and training methods as applied to training in Diplomacy. For details on the exact parameters, see Appendix \ref{app:params}.
\subsection{Data Generation and Training}

We implement deep Nash value iteration following the same considerations as deep Q-learning. Multiple data generation workers run self-play games in parallel. The policy and value targets on each turn along with the game state they correspond to are collected in a shared experience replay buffer and sampled by a set of training workers to train the networks. The policy proposal and value networks used by the data generation workers are only periodically updated from those in training.

During self play, each agent plays according to its part of the computed approximate-NE policy $\sigma$ with probability $1 - \eps$, and with probability $\eps$ plays a uniformly random action from the $N_c$ actions obtained from the policy proposal network.
We also use larger values of $\eps$ on the first two turns to quickly introduce diversity from the opening state, similar to the use of larger early exploration settings in other work, for example AlphaZero~\cite{silver2017mastering,silver2018general}. Detailed parameters are in Appendix \ref{app:params}.

\begin{algorithm}
\caption{Approximated Double Oracle}

\begin{algorithmic}[1]
\small
\Function{FindEquilibriumWithDO}{$s$, $N_c$, $N_p$, $N_{iters}$, $\eps$} 
    \State $\myvector{A} \gets $ \Call{SampleActionsFromProposalNet}{s, $N_c$}
    \State $\myvector{\sigma} \gets$ \Call{ComputeEquilibrium}{s, $\myvector{A}$}
    \For{$t=(1..N_{iters})$}
      \State $\textrm{modified} \gets 0$
      \For{$i=(1..N_{players})$}
        \State $\hat{A}_i \gets $ \Call{GenerateActions}{$s, A_i, N_p$} \Comment{Uniform sampling or local modification}
        \State $v_i \gets \mathbb{E}_{\myvector{a} \sim \myvector{\sigma}} \left[ V_i(f(s, \myvector{a}); \theta_v) \right]$
        \State $a_i^*, v_i^* \gets \argmax\limits_{a_i \in \hat{A}_i} \mathbb{E}_{\myvector{a_{-i}} \sim \myvector{\sigma_{-i}}} \left[ V_i(f(s, \myvector{a}); \theta_v) \right]$
        \If {$v_i^* - v_i > \eps$}
          \State $A_i \gets A_i \bigcup \{a_i^*\}$
          \State $\myvector{\sigma} \gets$ \Call{ComputeEquilibrium}{s, $\myvector{A}$}
          \State $\textrm{modified} \gets 1$
        \EndIf
      \EndFor
      \If {$!\textrm{modified}$}
        \State Break
      \EndIf
    \EndFor
    \Return $\myvector{\sigma}$
\EndFunction
\newline
\Function{GenerateActionsLocalModification}{$s$, $A_{base}$, $N_p$} 
    \State $A \gets \{\} $
    \While{$|A| < N_p$}
      \State $a \gets $ \Call{SampleAction}{$A_{base}$}
      \State $loc \gets $ \Call{SampleMapLocation}{}()
      \State $clique \gets $ \Call{GetNeighbours}{$loc$}
      \State $a' \gets $ \Call{RandomCoordinatedModification}{$a$, $clique$}
      \State $A \gets A \bigcup \{a'\}$
    \EndWhile
    \Return $A$

\EndFunction

\end{algorithmic}
\label{alg:rl_loop}
\end{algorithm}

\subsection{Model Architecture}
\label{sec:architecture}

Our model closely follows the encoder-decoder architecture of \cite{gray2020human}.
We use the policy head to model the policy proposal distribution and the value head to predict the expected values of the players.

We found several adjustments to be effective for our use in deep Nash-based RL. Firstly, we removed all dropout layers, as we found this to improve training stability. Secondly, we replaced the graph convolution-based encoder with a simpler transformer encoder, improving model quality and removing the need to hardwire the adjacency matrix of the Diplomacy map into the architecture. Lastly, we also split the network into separate policy and value networks. Since our regret minimization procedure is essentially only a 1-ply search, the states on which we need to evaluate policy (the current state) and value (the states after one turn) are disjoint, so there is no gain to computing both policy and value in one network query, and we found splitting the networks improved the quality and effective capacity of the models. 
We measure the impact of these choices in Table~\ref{tab:res_arch} and describe the architecture in more detail in Appendix~\ref{app:architecture}.




\subsection{Exploitability Testing}
\label{sec:exploitability}
To obtain a metric of the exploitablity of our final agents, we also train additional exploiter agents that rather than attempting to play an equilibrium, instead hold the exploited agent fixed and attempt to converge to a best response. We compare two approaches.

In the first approach, we reimplement the same actor-critic method with sampled entropy regularization previously used in~\cite{gray2020human} for 7-player no-press Diplomacy and apply it to 2-player FvA. In this approach, the other agent is treated as a black box, reducing the problem to just a standard Markov Decision Process to which we can apply classical RL techniques.
Namely, we use Impala~\cite{espeholt2018impala}, an asynchronous version of actor-critic algorithm.
Since search agents may take several seconds to act, \cite{gray2020human} first trained a policy network to copy the behavior of the search agent and then applied RL to exploit this proxy agent. The proxy agent acts much faster, which enabled generating data faster.
In contrast, we optimized the training loop so that it was possible to train to exploit the search agent directly without first generating a proxy policy network agent.

In the second approach, we begin with the policy and value models of the exploited agent and resume deep Nash value iteration, except with a single simple modification, described in Appendix \ref{app:exploitability}, that results in learning a best response rather than an equilibrium.

\section{Results}
Diplomacy features two main challenges for AI research: its large combinatorial action space and its multi-agent (non-2p0s) nature. \botname{} is intended to address the former challenge. To measure its effectiveness on just the action-exploration aspect, we first present results in a 2p0s variant of Diplomacy called France vs. Austria (FvA). We show \botname{} decisively beats top humans in FvA.


Next, we present results for 7-player no-press Diplomacy, which has been a domain for much prior research~\cite{paquette2019no,anthony2020learning,gray2020human}. We train a \botname{} agent from scratch with no human data and test it against pre-trained models from previous work~\cite{paquette2019no,gray2020human}\footnote{See Appendix~\ref{app:params} for the details.}. In the 6v1 setting, (six \botname{} agents playing with one copy of another bot) \botname{} wins overwhelmingly against all other agents. However, in the 1v6 setting (one \botname{} agent playing with six copies of another bot), \botname{} does slightly worse than previous search agents that leveraged human data. Our results suggest that \botname{} converges to strong play within an equilibrium (i.e., a ``metagame'') that is very different from how humans play.
Finally, we present results showing that adding our RL algorithm on top of a human-bootstrapped model clearly beats prior bots in both the 1v6 and 6v1 settings.




\subsection{Results in 2-player France vs. Austria}
\label{sec:resultsfva}


\begin{table}[h]
\begin{center}
    \begin{tabular}{l|r|r|r}
         & \textbf{Austria} & \textbf{France} & \textbf{Total} \\
        \toprule
        Average score (+variance reduction) & $89\% \scriptstyle\pm 5\%$ & $66\% \scriptstyle\pm ~7\%$ & $78\% \scriptstyle\pm 4\%$ \\
        Raw average score & $100\% \scriptstyle\pm 0\%$ & $73\% \scriptstyle\pm 12\%$ & $87\% \scriptstyle\pm 6\%$ \\
        \# games & $13$ & $13$ & $26$ \\
        \bottomrule
    \end{tabular}
\end{center}
\caption{
\small
Head-to-head score of \botFvA{} versus top FvA players on \url{webdiplomacy.net}. The $\pm$ shows one standard error. Variance reduction (Appendix \ref{sec:variance_reduction}) was used to compute average scores from raw average scores. Note that for raw Austria score, because DORA managed to win \emph{all} 13 games, anomalously the sample standard deviation is technically 0. The true standard deviation is of course larger but cannot be estimated.
}
\label{tab:res_human}
\end{table}

In 2-player FvA, we tested our final \botFvA{} agent against multiple top players of this variant, inviting them to play a series of games against our agent. All players were allowed to play an arbitrary number of practice games against the agent before playing official counted games. Prior to running the games, we also implemented a simple variance reduction method, described in Appendix~\ref{sec:variance_reduction}.

In Table \ref{tab:res_human} we report both the raw and variance-reduced average scores. DORA defeats top human players by a large and highly significant margin, with an average score of 78\% after variance reduction. It also wins significantly more than half of games as France, the side considered much harder to play\footnote{Between agents of similar strength we often observe France winning only about a third of games vs Austria.}. These results show DORA achieves a level of play in FvA above that of top humans.

We ran several ablations to loosely estimate the relative importance of the various components contributing to our final agent for France vs Austria.

\paragraph{GraphConv vs Transformers}
\begin{table}[h]
\begin{center}
    \begin{tabular}{l|r|r }
         Architecture & Final Elo & Params \\
        \toprule
        GraphConv~\cite{gray2020human} & $ 898 \pm 13 $& 46.8M \\
        \midrule
        TransformerEnc 5x192 & $1041 \pm~~ 7$ & 9.3M \\
        TransformerEnc 10x224 & $1061 \pm 12$ & 12.4M \\
        TransformerEnc 10x224 (Separate) & $1123 \pm 13$ & 24.8M \\
        \bottomrule
    \end{tabular}
\end{center}
\caption{
\small
Effect in FvA of the architecture on final strength after RL, and number of parameters. GraphConv is the architecture achieving SOTA in supervised learning from~\cite{gray2020human}, and similar to other past work~\cite{paquette2019no,anthony2020learning}. TransformerEnc $b\,$x$f\,$ replaces the graph-conv encoder with a transformer encoder with $b$ blocks and $f$ feature channels. "Separate" uses separate value and policy networks instead of a shared network with two heads. Elos are based on 100 games against each of several internal baseline agents, including the agents in Table~\ref{tab:do_inference}.}
\label{tab:res_arch}
\end{table}

\paragraph{Double Oracle at Training Time}
\begin{table}[h]
\begin{center}
    \begin{tabular}{l|r|r }
        DO Method During RL Training  & Final Elo \\
        \toprule
        No DO & $ 927 \pm 13$  \\
        DO, uniform random sampling & $ 956 \pm 13$  \\
        DO, local modification sampling & $1023 \pm 13$ \\
        \bottomrule
    \end{tabular}
\end{center}
\caption{
\small
Effect of DO methods during training on final FvA agent strength. Local modification sampling improves results greatly, whereas uniform random sampling is relatively ineffective. Elos $\pm$ one standard error were computed based on 100 games against each of several baseline agents, including the agents in Table~\ref{tab:do_inference}.
}
\label{tab:selfplay_do}
\end{table}

See Table~\ref{tab:res_arch}. Although not a primary focus of this paper, we found that at least for use in RL, changing our model architecture to use a transformer encoder rather than a graph-convolution encoder in Section \ref{sec:architecture} resulted in significant gains over prior neural network architectures in Diplomacy~\cite{paquette2019no,anthony2020learning,gray2020human}, while using fewer parameters and still being similarly fast or faster in performance. Enlarging the model and splitting the network led to further gains.

See Table~\ref{tab:selfplay_do}. We demonstrate the effectiveness of our DO method in finding improved actions during training. DO's ability to discover new strong actions for the policy proposal network to learn makes a big difference in the final agent. Local modification sampling is effective, whereas random sampling is relatively ineffective.

\paragraph{Double Oracle at Inference Time}

\begin{table}
    \centering
    \begin{tabular}{l|rrr}
\multicolumn{1}{r}{Model used by both players $\rightarrow$ } & Model1 & Model2 & Model3 \\
DO Method used by measured player $\downarrow$ &  &  &  \\
\toprule
None & $0.5 \scriptstyle\pm 0$ & $0.5 \scriptstyle\pm 0$ & $0.5 \scriptstyle\pm 0$ \\
\midrule
DO, uniform random sampling & $0.75 \scriptstyle\pm 0.02$ & $0.48 \scriptstyle\pm 0.02$ & $0.47 \scriptstyle\pm 0.02$ \\
DO, local modification sampling & $0.89 \scriptstyle\pm 0.02$ & $0.59 \scriptstyle\pm 0.02$ & $0.56 \scriptstyle\pm 0.02$ \\
\bottomrule
\end{tabular}
\caption{
\small
Effect of using DO at inference time in FvA. Each entry is the average score when both players play the column agent, but one player uses a DO variant at inference time. Model1-Model3 are models of increasing quality developed over the course of our research, with the major differences being that Model1 uses a trained value network with a fixed proposal network, Model2 uses GraphConv encoder, and Model3 uses a small Transformer encoder. Stronger agents require better sampling methods to find candidate actions that help.
}
\label{tab:do_inference}
\end{table}

See Table~\ref{tab:do_inference}. We also demonstrate the effectiveness of our DO method in finding improved candidate actions at inference time. Uniform random sampling among legal actions to generate the action pool for finding best responses only improves the weakest of our baseline agent at inference time, due to being ineffective at finding better actions within the huge action space, whereas local modification sampling improves all baselines at inference time.

\subsubsection{Exploitability}
\begin{table}[h]
\begin{center}
    \begin{tabular}{l|r|r}
         & \textbf{vs DORA} & \textbf{vs DORA w/ inference-time DO} \\
        \toprule
        Actor-Critic PG~\cite{gray2020human} & $77.0\% \scriptstyle\pm 0.6\%$ & - \\
        Deep Nash-Derived Exploiter & $90.5\%  \scriptstyle\pm 0.4\%$ & $85.8\%  \scriptstyle\pm 0.5\%$  \\
        Deep Nash-Derived Exploiter, values only & $65.7\% \scriptstyle\pm 0.6\%$ & $60.5\%  \scriptstyle\pm 0.7\%$ \\
        \bottomrule
    \end{tabular}
    
\end{center}
\caption{
\small
Win rates for exploiter bots trained with different techniques against our best FvA agent with and without inference-time DO. The $\pm$ shows one standard error. Actor-Critic PG vs DORA with inference-time DO was not run due to computational cost.
}
\label{tab:exploit}
\end{table}


We implemented and tested both algorithms from Section \ref{sec:exploitability} to estimate a lower bound on the exploitability of our agents. We also test a variant where we adapt deep Nash value iteration for exploitation just as in the second method, but disable policy learning - the value network learns to predict states of higher exploitability, but the policy proposal network is fixed and does not learn.

The results against our best agent are in Table~\ref{tab:exploit}. We find that despite being of superhuman strength, the agent is still relatively exploitable. In a 2p0s large-action-space setting, this is not necessarily surprising. For example in Go, another game with a moderately large action space, there exist gaps of more than 1000 Elo (nearly 100\% winrate) between different agents, all of which are superhuman, even with no explicit training for exploitation~\cite{silver2016mastering}. Moreover, in imperfect-information games such as Diplomacy, exploiting a fixed agent is much easier than training an agent with low exploitability, since the exploiting agent does not need to precisely balance the probabilities of its actions.

Nonetheless, we also observe that our adaptation of the deep Nash value iteration method greatly outperforms the simpler actor-critic method from~\cite{gray2020human}, and that both value learning and DO-based policy learning are important for its performance, with value learning improving over the baseline from $50\%$ to $60.5\%$, and policy learning improving further to $85.8\%$ winrate over \botFvA{}. 

We also observe that applying DO at inference time reduces the exploitability, from $90.5\%$ using our strongest exploiter agent to $85.8\%$. Even at inference time, DO continues to find and suggest new effective actions that would otherwise be overlooked by the policy proposal network.

\subsection{Results in 7-player No-Press Diplomacy}

We also applied our methods to 7-player no-press Diplomacy, training a \botname{} agent entirely from scratch with no human data in this more-challenging domain. For comparison, we also trained an agent via deep Nash value iteration beginning from policy and value networks first trained on human games.
Our best results were achieved when we forced the agent to use the human-initialized policy for action proposal during training games, never updating it, and also stopping the training early - see additional ablations in Appendix~\ref{sec:extra_7p_results}). Since we do not update the action proposal policy used in training games, we also do not apply DO for this agent, and so we refer to this agent as~\botHumanInit{} ("\emph{d}eep \emph{N}ash \emph{v}alue \emph{i}teration, \emph{n}o \emph{p}roposal \emph{u}pdate").
The results are reported in Table~\ref{tab:res_7p}. Note that as these are 7-player games, the expected score if the agents perform equally well is $1/7 \sim 14.3\%$.

\begin{table}[h]
\begin{center}
    \begin{tabular}{l|r|r|r|r}
        1x $\downarrow$ vs 6x $\rightarrow$ & DipNet~\cite{paquette2019no}& SearchBot~\cite{gray2020human} & \botname{} & \botHumanInit{} \\
        \toprule
        DipNet~\cite{paquette2019no} & - & $0.8\% \scriptstyle\pm 0.4\%$ & $0.0\% \scriptstyle\pm 0.0\%$ & $0.1\% \scriptstyle\pm 0.0\%$ \\
        SearchBot~\cite{gray2020human} & $49.4\% \scriptstyle\pm 2.6\%$ & - & $1.1\% \scriptstyle\pm 0.4\%$ & $0.5\% \scriptstyle\pm 0.2\%$ \\
        \botname{} & $22.8\% \scriptstyle\pm 2.2\%$ & $11.0\% \scriptstyle\pm 1.5\%$  & - & $2.2\% \scriptstyle\pm 0.4\% $  \\
        \botHumanInit{}  & $45.6\% \scriptstyle\pm 2.6\%$ &  $36.3\% \scriptstyle\pm 2.4\%$ & $3.2\%  \scriptstyle\pm 0.7\% $ & - \\
        \midrule
        DipNet-Transf & $23.4\% \scriptstyle\pm 2.2\%$ & $2.1\% \scriptstyle\pm 0.7\%$ & $0.0\% \scriptstyle\pm 0.0\%$ & $0.3\% \scriptstyle\pm 0.1\%$ \\
        SearchBot-Transf & $48.1\% \scriptstyle\pm 2.6\%$ & $13.9\% \scriptstyle \pm 1.7\%$ &  $0.5\% \scriptstyle\pm 0.3\%$ & $0.9\% \scriptstyle\pm 0.2\%$ \\
        \bottomrule
    \end{tabular} 
\end{center}
\caption{
\small
SoS scores of various agents playing against 6 copies of another agent. The $\pm$ shows one standard error. Note that equal performance is 1/7 $\approx$ 14.3\%. The non-human DORA performs very well as a 6x opponent, while performing much worse on the 1x side with 6 human-like opponents.
The last 2 rows are equivalent to first 2 rows, but are retrained with the same architecture as self-play agents, i.e., TransformerEnc~5x192, showing the effect of architecture on these baselines separately from our deep Nash RL methods.
}
\label{tab:res_7p}
\end{table}

\begin{table}[h]
\begin{center}
    \begin{tabular}{l|r|r}
        1x $\downarrow$ vs 6x $\rightarrow$ & \botname{} & \botname{} (alternate run) \\
        \toprule
        \botname{} & - & $7.5\% \scriptstyle\pm 1.0\%$ \\
        \botname{} (alternate run) & $3.2\% \scriptstyle\pm 0.7\%$ & - \\
        
        \bottomrule
    \end{tabular} 
\end{center}
\caption{
\small
SoS scores between two runs of DORA, 1 agent versus 6 copies of the other. The $\pm$ shows one standard error. Note that equal performance would be 1/7 $\approx$ 14.3\%.
}
\label{tab:res_doraalt}
\end{table}

Unlike in 2p0s settings, in a competitive multi-agent game, particularly one such as Diplomacy where agents must temporarily ally and coordinate to make progress, there may be multiple different equilibria that are mutually incompatible~\cite{luce1989games}. An agent playing optimally under one equilibrium may perform very poorly in a pool of agents playing a different equilibrium.

For example in Diplomacy, it may be mutually benefit two agents to ally to attack a third, but in such cases players must often coordinate on a choice, for example on which unit attacks and which unit supports, or on which region to attack. Since an order and a support must exactly match to be effective, each choice may be a stable equilibrium on its own. But an agent that plays one choice in a population of agents that always play another in that situation will fail to match and will do poorly.

Our results are consistent with \botname{} converging roughly to an equilibrium that is very different than the one played by humans, but playing well within that equilibrium. \botname{} achieves only about an 11\% winrate in a 1-vs-6 setting against one of the strongest prior human-data-based agents, SearchBot~\cite{gray2020human} (tying would be a score of 1/7 $\approx$ 14.3\%). However, SearchBot fares even worse with only a 1.1\% winrate in a 1-vs-6 setting against \botname{}.

Against a second DORA trained with a different random seed (Table~\ref{tab:res_doraalt}), each DORA scores only about 3-8\% against the other 1-vs-6. This suggests that different DORA runs may converge to different equilibria from each other.


Our results with \botHumanInit{} are also consistent with reinforcement learning converging to inconsistent equilibria. \botHumanInit{} soundly defeats SearchBot, the prior SOTA, winning 36.3\% of games 1-vs-6, while SearchBot wins only 0.5\% in the reverse matchup. Starting from a human-like policy, deep Nash value iteration massively improves the strength of the agent. Yet \botHumanInit{} still does extremely poorly vs \botname{}, and vice versa. Our results seem to suggest that in 7-player no-press Diplomacy, deep Nash value iteration converges to an equilibrium in which outside agents cannot compete, even agents trained independently by the same method, but that which equilibrium is found depends on initialization, with some equilibria performing far better versus human-like players than others. See also Appendix~\ref{sec:extra_7p_results} for additional ablations and experiments. 

\section{Conclusions and Future Work}

Diplomacy has been a benchmark for AI research for two main reasons: its massive combinatorial action space and its multi-agent nature. We show that \botname{} achieves superhuman performance in 2-player Diplomacy without any human data or reward shaping, despite the combinatorial action space and imperfect information. \botname{} therefore effectively addresses the first challenge of Diplomacy.

Additionally, we show that \botname{} in full-scale 7-player no-press Diplomacy approximates an equilibrium that differs radically from human-data-based agents. When one human-data-based agent plays against a population of \botname{} agents, the one agent loses by a wide margin. Similarly, one \botname{} agent against a population of human-data-based agents does not exceed an average level of performance. These 7-player results provide evidence that self-play from scratch may be insufficient to achieve superhuman performance in Diplomacy, unlike other multiplayer games such as 6-player poker~\cite{brown2019superhuman}. Furthermore, the results suggest that there may be a wide space of equilibria, with some differing dramatically from human conventions. \botname{} provides the first competitive technique for exploring this wider space of equilibria without being constrained to human priors.

\newpage



\bibliography{references}

\begin{thebibliography}{10}

\bibitem{lukas2020continuous}
Lukáš Adam, Rostislav Horčík, Tomáš Kasl, and Tomáš Kroupa.
\newblock Double oracle algorithm for computing equilibria in continuous games.
\newblock 09 2020.

\bibitem{ghostratings}
Thomas Anthony.
\newblock Ghost-ratings, 2020.

\bibitem{anthony2020learning}
Thomas Anthony, Tom Eccles, Andrea Tacchetti, J{\'a}nos Kram{\'a}r, Ian Gemp,
  Thomas~C Hudson, Nicolas Porcel, Marc Lanctot, Julien P{\'e}rolat, Richard
  Everett, et~al.
\newblock Learning to play no-press diplomacy with best response policy
  iteration.
\newblock {\em arXiv preprint arXiv:2006.04635}, 2020.

\bibitem{berner2019dota}
Christopher Berner, Greg Brockman, Brooke Chan, Vicki Cheung, Przemys{\l}aw
  D{\k{e}}biak, Christy Dennison, David Farhi, Quirin Fischer, Shariq Hashme,
  Chris Hesse, et~al.
\newblock Dota 2 with large scale deep reinforcement learning.
\newblock {\em arXiv preprint arXiv:1912.06680}, 2019.

\bibitem{blackwell1956analog}
David Blackwell et~al.
\newblock An analog of the minimax theorem for vector payoffs.
\newblock {\em Pacific Journal of Mathematics}, 6(1):1--8, 1956.

\bibitem{brown2020combining}
Noam Brown, Anton Bakhtin, Adam Lerer, and Qucheng Gong.
\newblock Combining deep reinforcement learning and search for
  imperfect-information games.
\newblock {\em arXiv preprint arXiv:2007.13544}, 2020.

\bibitem{brown2015simultaneous}
Noam Brown and Tuomas Sandholm.
\newblock Simultaneous abstraction and equilibrium finding in games.
\newblock In {\em Twenty-Fourth International Joint Conference on Artificial
  Intelligence}, 2015.

\bibitem{brown2019solving}
Noam Brown and Tuomas Sandholm.
\newblock Solving imperfect-information games via discounted regret
  minimization.
\newblock In {\em Proceedings of the AAAI Conference on Artificial
  Intelligence}, volume~33, pages 1829--1836, 2019.

\bibitem{brown2019superhuman}
Noam Brown and Tuomas Sandholm.
\newblock Superhuman {A}{I} for multiplayer poker.
\newblock {\em Science}, page eaay2400, 2019.

\bibitem{espeholt2018impala}
Lasse Espeholt, Hubert Soyer, Remi Munos, Karen Simonyan, Volodymir Mnih, Tom
  Ward, Yotam Doron, Vlad Firoiu, Tim Harley, Iain Dunning, et~al.
\newblock Impala: Scalable distributed deep-rl with importance weighted
  actor-learner architectures.
\newblock {\em arXiv preprint arXiv:1802.01561}, 2018.

\bibitem{gray2020human}
Jonathan Gray, Adam Lerer, Anton Bakhtin, and Noam Brown.
\newblock Human-level performance in no-press diplomacy via equilibrium search.
\newblock {\em arXiv preprint arXiv:2010.02923}, 2020.

\bibitem{hannan1957approximation}
James Hannan.
\newblock Approximation to bayes risk in repeated play.
\newblock {\em Contributions to the Theory of Games}, 3:97--139, 1957.

\bibitem{hart2000simple}
Sergiu Hart and Andreu Mas-Colell.
\newblock A simple adaptive procedure leading to correlated equilibrium.
\newblock {\em Econometrica}, 68(5):1127--1150, 2000.

\bibitem{hu2003nash}
Junling Hu and Michael~P Wellman.
\newblock Nash q-learning for general-sum stochastic games.
\newblock {\em Journal of machine learning research}, 4(Nov):1039--1069, 2003.

\bibitem{jain2011double}
Manish Jain, Dmytro Korzhyk, Ond{\v{r}}ej Van{\v{e}}k, Vincent Conitzer, Michal
  P{\v{e}}chou{\v{c}}ek, and Milind Tambe.
\newblock A double oracle algorithm for zero-sum security games on graphs.
\newblock In {\em The 10th International Conference on Autonomous Agents and
  Multiagent Systems-Volume 1}, pages 327--334, 2011.

\bibitem{kingma2014adam}
Diederik~P Kingma and Jimmy Ba.
\newblock Adam: A method for stochastic optimization.
\newblock {\em arXiv preprint arXiv:1412.6980}, 2014.

\bibitem{lanctot2009monte}
Marc Lanctot, Kevin Waugh, Martin Zinkevich, and Michael Bowling.
\newblock Monte carlo sampling for regret minimization in extensive games.
\newblock In {\em Advances in neural information processing systems}, pages
  1078--1086, 2009.

\bibitem{lanctot2017psro}
Marc Lanctot, Vinicius Zambaldi, Audrunas Gruslys, Angeliki Lazaridou, Karl
  Tuyls, Julien Pérolat, David Silver, and Thore Graepel.
\newblock A unified game-theoretic approach to multiagent reinforcement
  learning.
\newblock In {\em nternational Conference on Neural Information Processing
  Systems}, pages 4193--–4206, 2017.

\bibitem{littman1994markov}
Michael~L Littman.
\newblock Markov games as a framework for multi-agent reinforcement learning.
\newblock In {\em Machine learning proceedings 1994}, pages 157--163. Elsevier,
  1994.

\bibitem{luce1989games}
R~Duncan Luce and Howard Raiffa.
\newblock {\em Games and decisions: Introduction and critical survey}.
\newblock Courier Corporation, 1989.

\bibitem{mcaleer2020pipeline}
Stephen Mcaleer, JB~Lanier, Roy Fox, and Pierre Baldi.
\newblock Pipeline psro: A scalable approach for finding approximate nash
  equilibria in large games.
\newblock In H.~Larochelle, M.~Ranzato, R.~Hadsell, M.~F. Balcan, and H.~Lin,
  editors, {\em Advances in Neural Information Processing Systems}, volume~33,
  pages 20238--20248. Curran Associates, Inc., 2020.

\bibitem{mcmahan2003doubleoracle}
Brendan McMahan, Geoffrey Gordon, and Avrim Blum.
\newblock Planning in the presence of cost functions controlled by an
  adversary.
\newblock In {\em International conference on machine learning}, pages
  536–--543, 2003.

\bibitem{mnih2015human}
Volodymyr Mnih, Koray Kavukcuoglu, David Silver, Andrei~A Rusu, Joel Veness,
  Marc~G Bellemare, Alex Graves, Martin Riedmiller, Andreas~K Fidjeland, Georg
  Ostrovski, et~al.
\newblock Human-level control through deep reinforcement learning.
\newblock {\em Nature}, 518(7540):529--533, 2015.

\bibitem{paquette2019no}
Philip Paquette, Yuchen Lu, Seton~Steven Bocco, Max Smith, O-G Satya,
  Jonathan~K Kummerfeld, Joelle Pineau, Satinder Singh, and Aaron~C Courville.
\newblock No-press diplomacy: Modeling multi-agent gameplay.
\newblock In {\em Advances in Neural Information Processing Systems}, pages
  4474--4485, 2019.

\bibitem{paszke2019pytorch}
Adam Paszke, Sam Gross, Francisco Massa, Adam Lerer, James Bradbury, Gregory
  Chanan, Trevor Killeen, Zeming Lin, Natalia Gimelshein, Luca Antiga, et~al.
\newblock Pytorch: An imperative style, high-performance deep learning library.
\newblock In {\em Advances in neural information processing systems}, pages
  8026--8037, 2019.

\bibitem{shapley1953stochastic}
Lloyd~S Shapley.
\newblock Stochastic games.
\newblock {\em Proceedings of the national academy of sciences},
  39(10):1095--1100, 1953.

\bibitem{silver2016mastering}
David Silver, Aja Huang, Chris~J Maddison, Arthur Guez, Laurent Sifre, George
  Van Den~Driessche, Julian Schrittwieser, Ioannis Antonoglou, Veda
  Panneershelvam, Marc Lanctot, et~al.
\newblock Mastering the game of go with deep neural networks and tree search.
\newblock {\em Nature}, 529(7587):484, 2016.

\bibitem{silver2018general}
David Silver, Thomas Hubert, Julian Schrittwieser, Ioannis Antonoglou, Matthew
  Lai, Arthur Guez, Marc Lanctot, Laurent Sifre, Dharshan Kumaran, Thore
  Graepel, et~al.
\newblock A general reinforcement learning algorithm that masters chess, shogi,
  and go through self-play.
\newblock {\em Science}, 362(6419):1140--1144, 2018.

\bibitem{silver2017mastering}
David Silver, Julian Schrittwieser, Karen Simonyan, Ioannis Antonoglou, Aja
  Huang, Arthur Guez, Thomas Hubert, Lucas Baker, Matthew Lai, Adrian Bolton,
  et~al.
\newblock Mastering the game of go without human knowledge.
\newblock {\em Nature}, 550(7676):354, 2017.

\bibitem{syrgkanis2015fast}
Vasilis Syrgkanis, Alekh Agarwal, Haipeng Luo, and Robert~E Schapire.
\newblock Fast convergence of regularized learning in games.
\newblock In {\em Advances in Neural Information Processing Systems}, pages
  2989--2997, 2015.

\bibitem{van2020q}
Tom Van~de Wiele, David Warde-Farley, Andriy Mnih, and Volodymyr Mnih.
\newblock Q-learning in enormous action spaces via amortized approximate
  maximization.
\newblock {\em arXiv preprint arXiv:2001.08116}, 2020.

\bibitem{vinyals2019grandmaster}
Oriol Vinyals, Igor Babuschkin, Wojciech~M Czarnecki, Micha{\"e}l Mathieu,
  Andrew Dudzik, Junyoung Chung, David~H Choi, Richard Powell, Timo Ewalds,
  Petko Georgiev, et~al.
\newblock Grandmaster level in starcraft ii using multi-agent reinforcement
  learning.
\newblock {\em Nature}, 575(7782):350--354, 2019.

\end{thebibliography}
\bibliographystyle{plain}

\newpage

\section{Checklist}
\begin{enumerate}


\item For all authors...
\begin{enumerate}
  \item Do the main claims made in the abstract and introduction accurately reflect the paper's contributions and scope?\\
  \answerYes
  \item Did you describe the limitations of your work? \\
   \answerYes Particularly with regards to non-2p0s settings, we discuss limitations interlaced through the discussion of theoretical grounding and through discussion of our results in 7-player and the conclusion.
  \item Did you discuss any potential negative societal impacts of your work?\\
    \answerNA We do not expect near-term societal impacts from this work.
  \item Have you read the ethics review guidelines and ensured that your paper conforms to them?\\
    \answerYes
\end{enumerate}

\item If you are including theoretical results...
\begin{enumerate}
  \item Did you state the full set of assumptions of all theoretical results?\\
    \answerYes
	\item Did you include complete proofs of all theoretical results?\\
    \answerYes In Appendix~\ref{app:convergence} that also refers to prior work that proves the main result we rely on.
\end{enumerate}

\item If you ran experiments...
\begin{enumerate}
  \item Did you include the code, data, and instructions needed to reproduce the main experimental results (either in the supplemental material or as a URL)?\\
    \answerNo The experiments in the paper require an extensive code base that is challenging to check for any violation of the Double Blind principles. We plan to publish the repository with camera-ready version. 
  \item Did you specify all the training details (e.g., data splits, hyperparameters, how they were chosen)? \\
    \answerYes Detailed parameters are provided in Appendix~\ref{app:params}.
	\item Did you report error bars (e.g., with respect to the random seed after running experiments multiple times)? \\
    \answerYes Error bars are given on relevant measurements and statistics.
	\item Did you include the total amount of compute and the type of resources used (e.g., type of GPUs, internal cluster, or cloud provider)?\\
    \answerYes We report the requirements in Appendix~\ref{app:params}.
\end{enumerate}

\item If you are using existing assets (e.g., code, data, models) or curating/releasing new assets...
\begin{enumerate}
  \item If your work uses existing assets, did you cite the creators?\\
    \answerYes
  \item Did you mention the license of the assets?\\
    \answerYes We specify the license of the model and code we use in appendix~\ref{app:params}. We plan to release the code and the models under MIT license.
  \item Did you include any new assets either in the supplemental material or as a URL?\\
    \answerNo As explained in 3a we plan to release the model and the code after the Double Blind review ends.
  \item Did you discuss whether and how consent was obtained from people whose data you're using/curating?\\
    \answerYes We discuss data used in appendix~\ref{app:human_exps}.
  \item Did you discuss whether the data you are using/curating contains personally identifiable information or offensive content?\\
    \answerYes We do not use any data with personally identifiable information or offensive content (see appendix~\ref{app:human_exps}).
\end{enumerate}

\item If you used crowdsourcing or conducted research with human subjects...
\begin{enumerate}
  \item Did you include the full text of instructions given to participants and screenshots, if applicable?\\
    \answerYes We included the instructions given to participants in appendix~\ref{app:human_exps}.
  \item Did you describe any potential participant risks, with links to Institutional Review Board (IRB) approvals, if applicable?\\
    \answerNA
  \item Did you include the estimated hourly wage paid to participants and the total amount spent on participant compensation?\\
    \answerNA
\end{enumerate}

\end{enumerate}

\appendix
\newpage
\section{Neural Network Architecture}
\label{app:architecture}
In this appendix, we describe the neural network architecture used for our agents.

\begin{figure}[h!]
\centering
\includegraphics[width=14.1cm]{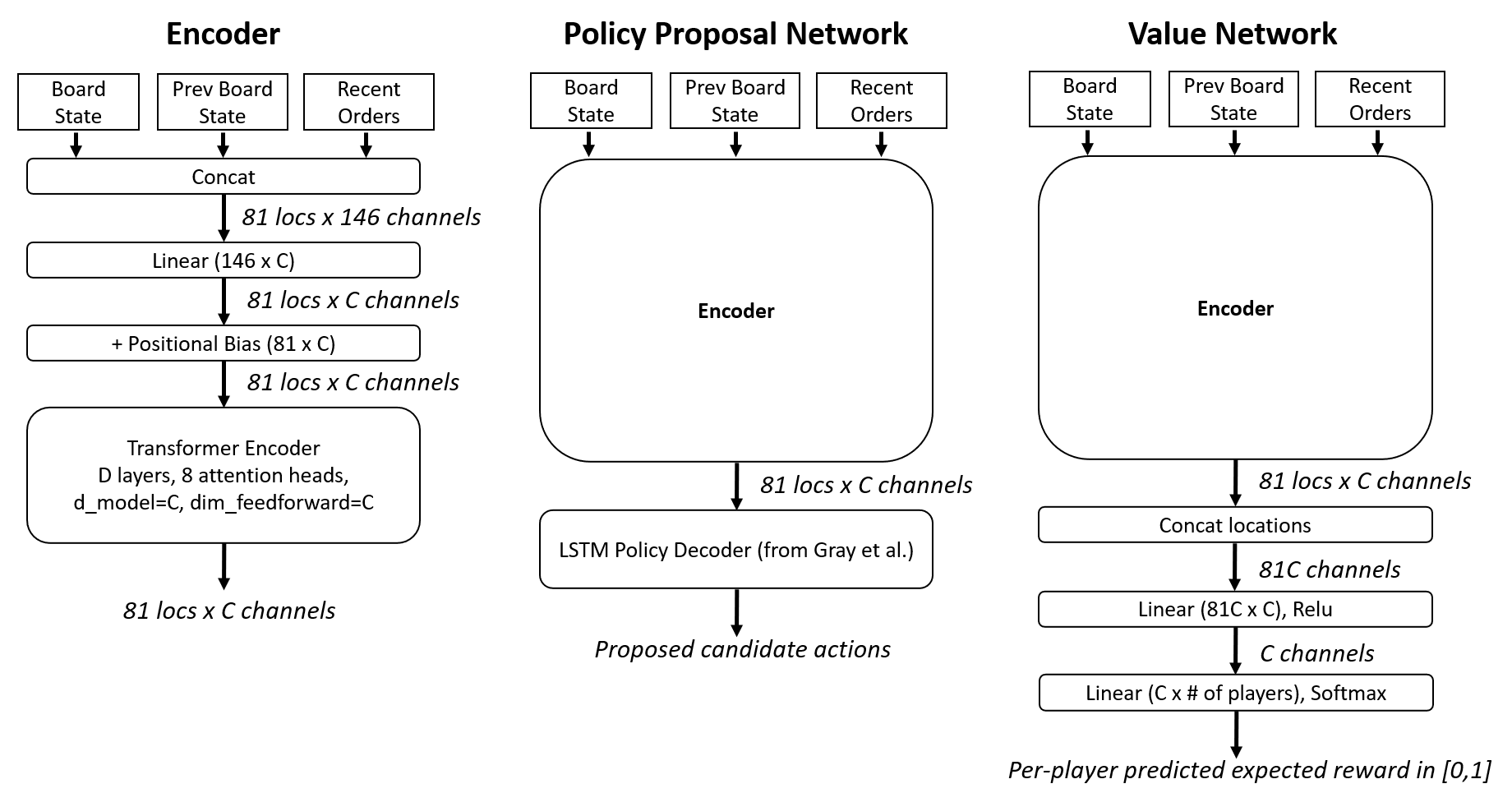}
\vspace{-0.2in}
\caption{\small Transformer encoder (left) used in both policy proposal network (center) and value network (right).}
\label{fig:arch}
\end{figure}

Our model architecture is shown in Figure \ref{fig:arch}. It is essentially identical to the architecture in \cite{gray2020human}, except that it replaces the specialized graph-convolution-based encoder with a much simpler transformer encoder, removes all dropout layers, and uses separate policy and value networks. Aside from the encoder, the other aspects of the architecture are the same, notably the LSTM policy decoder, which decodes orders through sequential attention over each successive location in the encoder output to produce an action.

The input to our new encoder is also identical to that of~\cite{gray2020human}, consisting of the same representation of the current board state, previous board state, and a recent order embedding. Rather than processing various parts of this input in two parallel trunks before combining them into a shared encoder trunk, we take the simpler approach of concatenating all features together at the start, resulting in 146 feature channels across each of 81 board locations (75 region + 6 coasts). We pass this through a linear layer, add pointwise a learnable per-position per-channel bias, and then pass this to a standard transformer encoder architecture.\footnote{For example, \url{https://pytorch.org/docs/stable/generated/torch.nn.TransformerEncoder.html}}

In our final agents, we use $(D,C) = (10,224)$ although in Section \ref{sec:resultsfva} we also present results for $(D,C) = (5,192)$ as well as for using a single combined network (i.e. shared encoder) for both policy and value.

\section{Theoretical Correctness of Nash Value Iteration}
\label{app:convergence}
In any 2p0s stochastic game with deterministic\footnote{We believe this result holds for games with non-deterministic transitions as well, but that does not allow for a direct reduction from Nash Q learning, so we do not provide a proof here.} transition function $f$, we show the Deep Nash Value Iteration (DNVI) procedure described in Section \ref{section:deepnashlearning} applied over a candidate set of \emph{all} legal actions converges to a value function $\hat{V}(s)$ that is the minimax value of each reachable state $s$, in the tabular setting. In other words, the use of a value function rather than a Q function does not affect the correctness.

To demonstrate this, we first show an equivalence between our procedure and Nash Q-learning in 2p0s games. At each step of DNVI, we update the value of $\myvector{V}(s)$ using the value computed by RM; since this stage game is 2p0s, RM converges to its unique minimax equilibrium value\cite{hart2000simple}. For each $(s, \myvector{a})$, $$\myvector{Q}(s,\myvector{a})= \myvector{r}(f(s,\myvector{a})) + \myvector{V}(f(s,\myvector{a})),$$ therefore each time $\myvector{V}$ is updated at state $s$ it is equivalent to applying a Nash-Q update at all $(s', \myvector{a})$ for which $f(s', \myvector{a})=s$.

\cite{hu2003nash} proves that Nash Q-learning converges to the Q values of a Nash equilibrium for the full game in the tabular setting under the following assumptions:
\begin{enumerate}
    \item Every one-step game has an adversarial equilibrium and this equilibrium is used in the update rule.
    \item All states are visited an infinite number of times.
    \item The learning rate $\alpha$ decays such that $\alpha \rightarrow 0$ but $\sum \alpha = \infty$.
\end{enumerate}
If these assumptions are satisfied, then $\myvector{V}(s)$ converges to the values of a Nash equilibrium.

Regarding assumption 1, in 2p0s all NEs are automatically adversarial since there is only one opponent (\emph{adversarial} equilibria are NEs that are additionally robust to joint deviations by opponents~\cite{hu2003nash}).

Assumption 2 is satisfied as long as $\forall a, P(a|s)>0$, which DNVI achieves via $\eps$-Nash exploration over the candidate set of all legal actions, which plays a random action with probability $\eps > 0$. 
Note that the modification we use in practice explores only actions that are likely under the current policy proposal net.

Assumption 3 concerns the learning rate schedule, which can be set arbitrarily. In the Deep RL setting we find in practice a fixed learning rate with the Adam~\cite{kingma2014adam} optimizer works well.

Thus, DNVI converges to a NE in a 2p0s stochastic game. In a 2p0s game all equilibria have identical (minimax) values. Therefore playing a minimax policy in each stage game based on these values is optimal.

We remind the reader that in Diplomacy the action space is too large to perform exact RM over the entire action space, so in practice we perform DNVI on a small subset of candidate actions proposed by the policy network and rely on Double Oracle to discover actions that should be added to this set.

\section{Variance Reduction in Diplomacy}\label{sec:variance_reduction}

We apply a simple form of variance reduction for our experiments playing with humans to help compensate for the relatively low sample size.

As in many simultaneous-action games, not uncommonly in Diplomacy there are situations that behave like a "matching pennies" subgame, where if player A plays action $a_1$, player B prefers to match it with an action $b_1$, and if A plays action $a_2$, player $B$ prefers to match it with action $b_2$, whereas player $A$'s preferences are the other way around, preferring the outcomes $(a_1,b_2)$ and $(a_2,b_1)$. Rock-paper-scissors-like or other kinds of simple subgames may also occur. In any of these cases, both players should randomize their strategies to reduce exploitability. See Table \ref{tab:pennies} for an example of a matching-pennies-like subgame.

\begin{table}[h]
\begin{center}
    \begin{tabular}{l|r|r}
         & $b_1$ & $b_2$ \\
        \hline
        $a_1$ & -1, 1 & 2,-2 \\
        $a_2$ & 1,-1 & -3, 3 \\
    \end{tabular}
\end{center}
\caption{Example payoff matrix for a matching-pennies-like game with nonuniform payoffs. In each cell, the payoff to the row player is listed first. }
\label{tab:pennies}
\end{table}

In such a subgame, once a player commits to playing a particular mixed strategy, then upon seeing the opponent's action, even without knowing the opponent's probability of playing that action, one can identify a significant component of variance due to the luck of randomization in one's \emph{own} mixed strategy that can be subtracted out. For example, in the example in Table \ref{tab:pennies} above if player A commits to a policy 60\% $a_1$ and 40\% $a_2$, then supposing player B is revealed to have chosen $b_2$, player A knows at that point that their expected value is $60\% \cdot 2$ (the payoff for $a_1$) $+ 40\% \cdot -3$ (the payoff for $a_2$). The deviation from the expected value due to whether player A \emph{actually} samples $a_1$ or $a_2$ on that round is now a matter of pure luck, i.e. just zero-mean variance that can be subtracted out.

In Diplomacy, we can apply this idea as well. When playing a game as agent $i$, on each turn we record the full distribution of our approximated NE $\sigma_i$ from which we sample the action $a_i$ that we submit that turn. Upon observing all other agents' actions $a_{-i}$, and letting $Q$ be the learned approximate Q-value function of our agent, we compute:
\[ \delta(a_i) = Q(s,(a_i,a_{-i})) - E_{a_i' \sim \pi_i} Q(s,(a_i',a_{-i})) \]
$\delta(a_i)$ measures how lucky we are to have chosen the specific $a_i$ that we did on that turn relative to the expected value over all actions $a_i'$ that we could have sampled from policy $\sigma_i$, given the observed actions of the other players. On each turn, we subtract this quantity from the final game win/loss reward of 1 or 0 that we ultimately observe. Since $E_{a_i} \delta(a_i) = 0$, doing this introduces no bias in expectation. So letting $\delta_t$ be the $\delta(a_i)$ value computed for turn $t$, and $R$ the total reward of the game, our final variance-adjusted result is 
\[ R - \sum_t \delta_t \] 

In the event that $Q$ is a good estimate of the true value of the state given the players' policies, then the $\delta_t$ values should be correlated with the final game outcome, so subtracting them should reduce the variance of the outcome, particularly in a 2-player setting such as FvA Diplomacy. In practice, we observe roughly a factor of 2 reduction in variance in informal test matches. As part of our internal practice, we also implemented and tested and committed to the use of this form of variance reduction prior to assembling the final human results.

\section{Implementation details}
\label{app:params}

We report optimizations for Double Oracle we use in practice, details of the pretraining procedure, modifications for 7p, and the hyper parameters used.

\subsection{Double Oracle optimizations}
We apply several modifications and approximations to DO to make it run faster in practice:

\begin{itemize}
\item We find a best response (line 8 of the algorithm~\ref{alg:rl_loop}) for only one player at a time. When $N_p$ is large, evaluating $N_p$ potential best responses is more expensive than computing a NE among the $N_c$ candidate actions ($N_{p} \times N_c$ value function calls versus $N_c \times N_c$).\footnote{When querying the value network is the dominating cost, the number of iterations needed to compute a NE is largely irrelevant because the value network results can be cached. For this reason, we consider the cost to be roughly $N_c \times N_c$ rather than $T \times N_c \times N_c$.} Recomputing the NE cheaply ensures each best response takes into account the newest action added by the opponent. Note, that for 7p caching is not as efficient due to large joint action space ($N_c^7$ instead of $N_c^2$), but computing BR still dominates the computations cost.
\item When finding a best response, we truncate the opponent's equilibrium policy, which spans up to $N_p$ actions, to only its $k$ highest-probability actions for some small $k$ and renormalize. This allows us to compute a best response with only $N_c * k$ value function calls rather than $N_c * N_p$ calls, while only introducing minimal error.
\item We cap the number of iterations of DO we execute.
\end{itemize}

\subsection{Pretraining}

We perform a short phase of pretraining before switching to deep Nash value iteration that improves the speed and stability of training. This phase differs from the main training phase in the following ways:

\begin{itemize}
\item Rather than query the policy proposal network and/or apply DO to obtain actions, we select $N_c$ candidate actions for each player uniformly at random among all legal actions.
\item Rather than training the value network to predict the 1-step value based on the computed equilibrium over the $N_c$ actions, we train it to directly predict the final game outcome.
\end{itemize}

By directly training on the final game outcome, this pre-training phase initializes the value network to a good starting point for the main training more quickly than bootstrapped value iteration. This phase also initializes the policy proposal network, which is trained to predict the output of regret matching on the random sampled actions, to a good high-entropy starting point - e.g. to predict a wide range of actions, mildly biased towards actions with high value. If instead a randomly initialized (untrained) policy proposal network were used to select actions, it may select a highly non-uniform initial distribution of candidate actions and require a lot of training for excluded actions to be rediscovered via DO.

\subsection{Extending to 7p}

We found a small number of additional implementation details were needed to handle 7-player no-press Diplomacy that were not needed for 2-player FvA. 

Firstly, exactly computing the value loss in equation \ref{eq:loss} is not feasible because the number of possible joint actions $\myvector{a'}$ scales with the power of the number of players, which for 7 players is too large. So we instead approximate the 1-step value via sampling.

Similarly, during DO, when computing the expected value of potential best responses, given a certain number of actions per opponent,  computing the value exactly scales with the power of the number of opponents, which for 6 opponents is too large. So again, we approximate by sampling.

Finally, in line with~\cite{gray2020human} we found that using Monte Carlo rollouts to compute state values improved the play compared to using a value function alone. Therefore, we use rollouts of depth 2 in all experiments at inference time. We do not use rollouts during training due to the computational cost.

\subsection{Hyper-parameters}

\subsubsection{Double Oracle}
We use slightly different parameters at training and inference time to speed up the data generation for training. See table~\ref{tab:do_hyperparams}.

\begin{table}[h]
\center
\begin{tabular}{lrr}
\toprule
 & Training & Inference \\
\midrule
Pool size ($N_p$)    & 1,000  & 10,000 \\
Max opponent action ($k$)    & 8  & 20 \\
Min value difference ($\eps$)    & 0.04  & 0.01 \\
Max iterations ($N_{iters}$)    & 6  & 16 \\
Pool recomputed after each iteration    & No  & Yes \\
\bottomrule
\end{tabular}
\caption{\small Hyper-parameter values used for Double Oracle for \botname.}
\label{tab:do_hyperparams}
\end{table}

\subsubsection{DORA training}

Details for training of FvA \botname{} bot are provided in table~\ref{tab:fva_dora_hyperparams}.
We use a few additional heuristics to facilate training that are explained below.

The training is bottlenecked by the data generation pipeline, and so we use only a few GPUs for training, but an order of magnitude more for data generation.
To make sure the training does not overfit when the generation speed is not enough, we throttle the training when the training to generation speeds ratio is above a threshold.
This number could be interpreted as a number of epochs over a fixed buffer.

Our final FvA \botname{} model is trained using 192 Nvidia V100 GPUs on an internal cluster, 4 used for the training and the rest is used for data generation. Both stages take around a week to complete. 

We made a number of changes for 7p training compared to FvA for computational efficiency reasons.
Each 7p game is 4-5 times longer and each equilibrium computation requires 3 times more operations due to the increased number of players.
Moreover, the cost of double oracle also increases many fold as the probability of finding a deviation for at least one player during a loop over players increases.
Therefore, for 7p \botname{} we add an additional pre-training stage where we train as normal, but without DO.
For speed reasons, we use a smaller transformer model for both the value and the policy proposal nets.

\begin{table}[h]
\center
\begin{tabular}{l|rr}
\toprule
 & FvA & 7p \\
\midrule
Learning rate & \multicolumn{2}{c}{$10^{-4}$}  \\
Gradient clipping &   \multicolumn{2}{c}{0.5} \\
Warmup updates & \multicolumn{2}{c}{10k} \\
Batch size & \multicolumn{2}{c}{1024} \\
Buffer size & \multicolumn{2}{c}{1,280,000}  \\
Max train/generation ratio & \multicolumn{2}{c}{6} \\
\midrule
Regret matching iterations    & \multicolumn{2}{c}{256}\\
Number of candidate actions ($N_c$)    & \multicolumn{2}{c}{50} \\
Max candidate actions per unit   & - & 6 \\
Number of sampled actions ($N_b$)    & \multicolumn{2}{c}{250} \\
Nash explore ($\varepsilon$)    & 0.1 & 0.1 \\
Nash explore, S1901M     & 0.8 & 0.3 \\
Nash explore, F1901M     & 0.5 & 0.2 \\
\bottomrule
\end{tabular}
\caption{\small Hyper-parameter values used to train \botname{} agents.}
\label{tab:fva_dora_hyperparams}
\end{table}

\subsubsection{Evaluation details}

We describe parameters used for agent evaluation in table~\ref{tab:res_7p}.

To compare against DipNet~\cite{paquette2019no} we use the original model checkpoint\footnote{DipNet SL from \url{https://github.com/diplomacy/research}. MIT License} and we sample from the policy with temperature $0.5$.
Similarly, to compare against SearchBot~\cite{gray2020human} agent we use the released checkpoint\footnote{blueprint from \url{https://github.com/facebookresearch/diplomacy_searchbot/releases/tag/1.0}. MIT License} and agent configuration\footnote{\url{https://github.com/facebookresearch/diplomacy_searchbot/blob/master/conf/common/agents/searchbot_02_fastbot.prototxt}}.
To make the comparison more fair, we used the same search parameters for \botname{} and \botHumanInit{} as for SearchBot (see table~\ref{tab:search_params_7p}).

\begin{table}[h]
\center
\begin{tabular}{l|r}
\toprule
Number candidate actions ($N_c$) & 50 \\
Max candidate actions per unit & 3.5 \\
Number CFR iterations & 256 \\
Policy sampling temperature for rollouts & 0.75 \\
Policy sampling top-p & 0.95 \\
Rollout length, move phases & 2 \\
\bottomrule
\end{tabular}
\caption{\small Parameters used for all 7p agents with search in table~\ref{tab:res_7p} in the main text.}
\label{tab:search_params_7p}
\end{table}

\subsubsection{Distributed training and data generation}
We use PyTorch~\cite{paszke2019pytorch} for distributed data parallel training and a custom framework for distributed data generation. We run several Python processes in parallel across multiple machines:

\begin{itemize}
    \item \textbf{Training processes.} All training processes run on one machine. Each is assigned a separate GPU and has a separate replay buffer from which it computes gradients, which are then broadcast and synchronized and across processes. One training process is also responsible for publishing new checkpoints for data generation workers to use and for collecting training statistics.
    \item \textbf{Data collection and rollout processes.} Each machine other than the training machine has a data collection process and many rollout processes. The data collection process gathers rollouts from rollout processes on the same machine via shared memory and sends them to the replay buffers via RPC. Rollout processes run the data generation loop. Each iteration of the loop consists of reading a new model checkpoint if available, stepping the game till the end, and sending results to the collection process. We run up to 8 rollout processes per GPU. Each process has its own copy of both the model and the environment.
    \item \textbf{Evaluation processes} run one-vs-six games of the current model checkpoint versus some fixed agent and collect running winrates for monitoring the run.
    \item \textbf{The metric collection process} receives metrics from other processes via RPC, aggregates and saves them. 
\end{itemize}

Our design was optimized to work on 8-GPU machines with 10 CPUs per GPU, such as Nvidia DGX-1 machines. Our main DORA run used 192 GPUs total, across 24 machines. In such a setting our setup achieves over 90\% average utilization for both GPU and CPU. 

\section{Adapting Deep Nash Value Iteration to Learn Best Responses}
\label{app:exploitability}

As mentioned in Section \ref{sec:exploitability}, our best exploiter agents resulted from adapting our main deep Nash value iteration to learn a best response instead of an equilibrium. 

To do so, we begin with the policy and value models of the exploited agent and resume deep Nash value iteration, except with a simple modification. On each turn of the game we first invoke the exploited agent to precompute its policy. Then the exploiting agent, when performing Nash learning, only performs RM for itself while the opponent samples only the precomputed policy. All other aspects of the architecture are identical.

Whereas normal RM approximates an NE, one-sided RM instead approximates a best response. With this one change, deep Nash value iteration, rather than training the networks to learn equilibrium policy and values, trains them to learn best-response exploitation and expected state-values assuming future best-response exploitation.

This second approach resembles the Sampled Best Response (SBR) exploiter from~\cite{anthony2020learning}, in that both methods use a policy model to sample candidate actions for the exploiter, and then approximate a 1-ply best response by directly maximizing against the exploited-agent's known policy with respect to a value model. However, unlike SBR, which uses policy and value models obtained via other methods such as supervised learning that may not be optimized for exploitation, our approach directly trains the policy to sample better exploitative candidate actions, and trains the value to also prefer states where exploitability on future turns will be high.

At test time, our exploiter agent takes the average of 3 samples of the exploited agent's reported average policy instead of only one to get a better estimate, since due to sampling noise or multiple equilibria, regret matching will sometimes return different final average policies. The exploited agent in test itself acts according to a 4th entirely independent sample, to ensure the exploiter can only optimize against the average and not the exact seed.

\section{Human data}\label{app:human_exps}

To conduct human experiments we informally reached out to top FvA Diplomacy players with a proposal to help with our research by playing against our agent.

The message contained the following instructions:

{
\tt
Hi [Person],

I'm AUTHORNAME, one of the developers of BOTNAME. We've put a new version of the bot online recently that we think might be superhuman, and we're hoping to do some testing against top humans to measure whether that's the case.

Would you be interested in playing ~10 games against the bot? You'd be able to play at your own pace. You're also welcome to play against the bot as many times as you'd like as practice before starting the "real" matches.

For the real matches, you could call them something like "[Person] vs BOTNAME 1", or something along those lines. You can launch them just like a normal game. We just ask that you not cancel any games and only put in a draw if it's clearly a stalemate.

These results (aggregated with other top players) would go into an academic paper. We could either use your real name or your username, or just report your FvA GhostRating, whichever you prefer.
}

Among the players that we contacted, 5 played matches against our bot. Those players were ranked 1st, 8th, 16th, 22nd, and 29th in FvA among all players on \url{webdiplomacy.net} according to the Ghost-Ratings~\cite{ghostratings} ranking system. None of the individual players had a positive win rate against the bot, though no single player played enough games against the bot to measure individual performance with statistical significance.

We did not collect any information regarding the games besides player's actions in the games played.
Thus, the data does not contain any personally identifiable information or offensive content.







\section{Action Exploration Example}
\label{sec:do_example}

\begin{figure}[!htb]
   \begin{minipage}{0.48\textwidth}
     \centering
     \includegraphics[width=.9\linewidth]{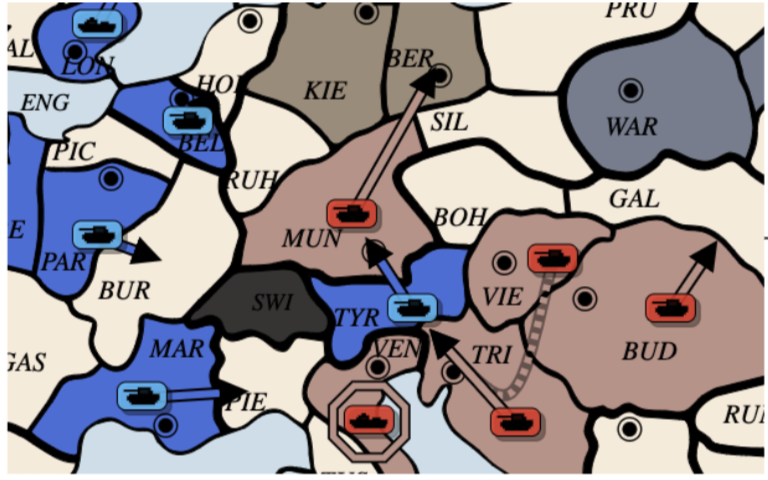}
   \end{minipage}\hfill
   \begin{minipage}{0.48\textwidth}
     \centering
     \includegraphics[width=.9\linewidth]{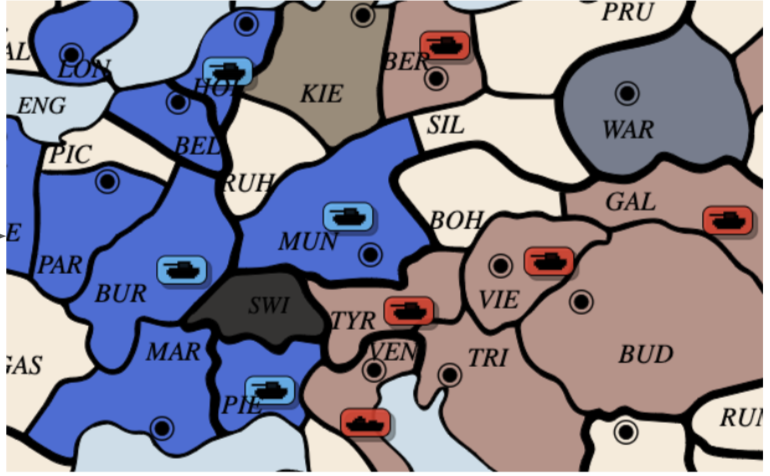}
   \end{minipage}
   \caption{The most probable actions for both players as predicted by the agent. The left figure shows the actions and the right figure shows the state after the actions are executed. France's army escapes to MUN.}\label{fig:do_real}
\end{figure}

We go into more detail regarding the example provided in Figure~\ref{fig:teaser} that motivated our double oracle approach.
This is an example of a situation from a real game played by our agent (France, blue) vs a human (Austria, red).
The agent has an army in Tyr next to the SCs of the opponent.
None of the actions that the agent considers for red could dislodge the army \emph{and} block its retreat further east.
Therefore, the estimated state value for blue in this position is high and one of the possible expected outcome is shown in Figure~\ref{fig:do_real}.

\begin{figure}[!htb]
   \begin{minipage}{0.48\textwidth}
     \centering
     \includegraphics[width=.9\linewidth]{figs/do_do1.png}
   \end{minipage}\hfill
   \begin{minipage}{0.48\textwidth}
     \centering
     \includegraphics[width=.9\linewidth]{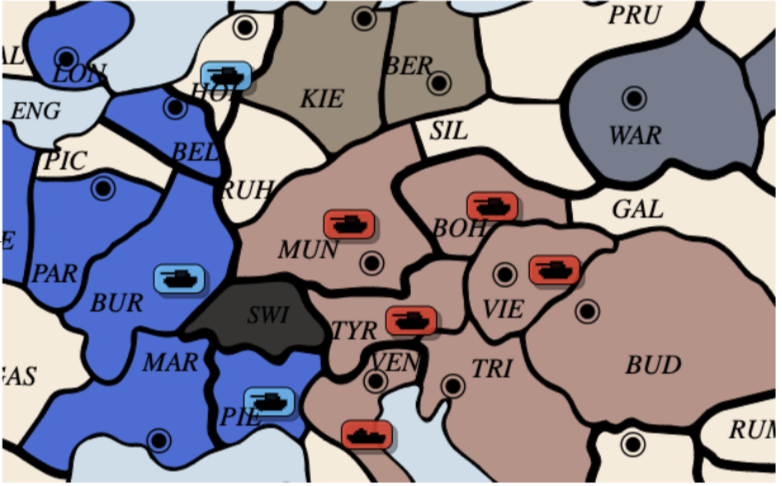}
   \end{minipage}
   \caption{The most probable action for France (blue) as predicted by the agent vs the best action for Austria (red) that the policy proposal network failed to find on its own. The left figure shows the actions and the right figure shows the state after the actions are executed. France's army is crushed and disbanded.}\label{fig:do_do}
\end{figure}

However, human players were quick to realize that there are two actions that can dislodge the blue army and force it to retreat to Pie (Figure~\ref{fig:do_do}, left).
Moreover, as the agent blocked its own retreat by moving Mar to Pie, the agent had to disband its army in Tyr~(Figure~\ref{fig:do_do}, right). Once the optimal action for red is added to the set of candidate actions, the probability of blue moving Mar to Pie and the expected score for blue in this state go down.

There are more than 4 million valid actions\footnote{Product of possible orders per location: MUN (17), SEV (6), VEN (8), VIE (18), TRI (18), BUD (16). Unit in SEV is not shown on the map.} or around 400,000 valid coordinated actions.
While it is still computationally feasible to run exact DO at inference time for this number of actions, adding even one more unit makes DO search infeasible.
Moreover, running DO at inference time only is not sufficient, as during optimal play an agent should know that this state has low value and so the actions leading to this state should be avoided.
Since the number of actions one can realistically evaluate during training is around 1,000, we have less than a 1\% chance of finding a BR action at training time.
Our approximate local modification approach allows us to dramatically reduce the search size and find good approximate BRs with relatively few actions.

Our final \botname{} agent finds red's action from Figure~\ref{fig:do_do} without using inference time DO, as the action is learned by the policy proposal network.

\newpage
\section{Additional Results for 7-player No-Press Diplomacy}\label{sec:extra_7p_results}

In this section we include more preliminary results exploring how initialization and training affect convergence to different equilibria for \botHumanInit{}.

Namely, we show in Table~\ref{tab:res_7p_extra} how performance versus human-like agents changes as self-play training progresses as well as the effect of freezing the action proposal network ("no policy update", "NPU").
As a reminder, \botHumanInit{} uses a training procedure identical to \botname{}, except a fixed blueprint from supervised learning on human games is used to propose actions throughout the training and hence no DO is used. The policy proposal net is still trained, but is only used at test time.
\botHumanInitWild{} is one step closer to \botname{}: it does update the policy proposal network during training on rollout workers, but does not use DO.
In general, both early stopping and NPU help the agent converge towards a strategy more effective against human-like models.

Additionally, we evaluated \botHumanInit{}-BP-Policy, an agent that uses the value function from~\botHumanInit{}, but the human blueprint policy proposal net during both training and test time. At all times this model performs worse than \botHumanInit{}. This suggests that a trained policy does increase the strength of the agent, even though using that policy within the RL loop makes the training too-easily diverge away from human-compatible strategies or equilibria. More research in the future may find better techniques to get the best of both worlds.

Finally, Table~\ref{tab:res_7p_extra_dora} provides additional data on the impact of DO for 7-player Diplomacy. We trained an agent from scratch for 300k updates without DO and then trained for 60k updates more with DO. This finetuning doubles the score versus SearchBot and also greatly increases it versus Dipnet.

\begin{table}[h]
\begin{center}
    \begin{tabular}{l|r|r|r}
        1x Agent & \# training updates & vs 6x DipNet~\cite{paquette2019no}& vs 6x SearchBot~\cite{gray2020human} \\
        \toprule
        DipNet~\cite{paquette2019no} & - & $0.8\% \scriptstyle\pm 0.4\%$  \\
        Transf &-& $23.4\% \scriptstyle\pm 2.2\%$ & $2.1\% \scriptstyle\pm 0.7\%$  \\
        SearchBot~\cite{gray2020human} & -& $49.4\% \scriptstyle\pm 2.6\%$ & -  \\
        Transf+Search &-& $48.1\% \scriptstyle\pm 2.6\%$ & $13.9\% \scriptstyle \pm 1.7\%$ \\
        \midrule
        \botname{} & 600k& $22.8\% \scriptstyle\pm 2.2\%$ & $11.0\% \scriptstyle\pm 1.5\%$    \\
        \midrule
        \botHumanInitWild{}  & 100k& $30.6\% \scriptstyle\pm 2.4\%$ &  $20.5\% \scriptstyle\pm 2.0\%$   \\
        \botHumanInitWild{}  & 300k& $25.3\% \scriptstyle\pm 2.3\%$ &  $18.6\% \scriptstyle\pm 2.0\%$   \\
        \midrule
        \botHumanInit{}  & 50k& $45.6\% \scriptstyle\pm 2.6\%$ &  $36.3\% \scriptstyle\pm 2.4\%$   \\
        \botHumanInit{}  & 100k& $41.4\% \scriptstyle\pm 2.5\%$ &  $34.3\% \scriptstyle\pm 2.3\%$   \\
        \botHumanInit{}  & 300k& $35.9\% \scriptstyle\pm 2.6\%$ &  $28.4\% \scriptstyle\pm 2.3\%$   \\
        \midrule
        \botHumanInit{}-BP-Policy  & 50k& - &  $25.7\% \scriptstyle\pm 2.1\%$   \\
        \botHumanInit{}-BP-Policy  & 100k& - &  $25.0\% \scriptstyle\pm 2.1\%$   \\
        \botHumanInit{}-BP-Policy  & 300k& - &  $22.9\% \scriptstyle\pm 2.0\%$   \\
        \bottomrule
    \end{tabular}
\end{center}
\caption{
\small
SoS scores of various agents playing against 6 copies of another agent. The $\pm$ shows one standard error. Note that equal performance is 1/7 $\approx$ 14.3\%. All our agents are based on TransformerEnc 5x192.
\vspace{-0.15in}
}
\label{tab:res_7p_extra}
\end{table}

\begin{table}[h]
\begin{center}
    \begin{tabular}{l|r|r|r}
        1x Agent & \# training updates & vs 6x DipNet~\cite{paquette2019no}& vs 6x SearchBot~\cite{gray2020human} \\
        \toprule
        \scratchDNLBot{}  & 300k& $13.6\% \scriptstyle\pm 1.8\%$ &  $3.5\% \scriptstyle\pm 0.9\%$   \\
        +finetune with DO  & +60k& $21.9\% \scriptstyle\pm 2.2\%$ &  $8.0\% \scriptstyle\pm 1.4\%$   \\
        \bottomrule
    \end{tabular}
\end{center}
\caption{
\small
Effect of using DO for training from scratch. Even training for a small fraction of time with DO increases the strength of an agent by a significant margin.
\vspace{-0.15in}
}
\label{tab:res_7p_extra_dora}
\end{table}

\end{document}